# Reliability, validity, and diagnostic evidence for multi-model LLM short-answer scoring

Chunyi Zhao and Chao Li

AIEd Governance

Emails: cccchunyi07@gmail.com; lichao@aegt.org

## Abstract

Large language models (LLMs) are increasingly used or proposed for educational scoring, but single-model and single-run evaluations provide limited evidence for assessment use. Short-answer scoring requires evidence about reliability, validity, severity, diagnostic value, and failure cases. This study evaluated repeated multi-model OCG-PRES guided LLM scoring for short-answer assessment. The analysis used 996 SciEntsBank responses. GPT, DeepSeek, and Qianwen each scored every response across three independent runs using five OCG-PRES dimensions: concept coverage, relation accuracy, reasoning completeness, contradiction control, and domain relevance. Scores were evaluated against official binary and five-category labels and compared with non-LLM baselines based on answer length, Jaccard keyword overlap, TF-IDF cosine similarity, and a combined traditional logistic model. Repeated-run reliability was high for all models, with ICC(3,k) = .977 for GPT, .992 for DeepSeek, and .981 for Qianwen. DeepSeek was the most stable across runs. GPT showed the strongest official-label alignment by AUC (.909), while Qianwen was stricter, with higher precision but lower recall under the fixed threshold = 3.0 rule. OCG-PRES scores followed expected diagnostic patterns across five official categories and outperformed all non-LLM baselines in AUC and F1. Repeated multi-model OCG-PRES scoring provides reliability, validity, and diagnostic evidence for LLM-assisted short-answer scoring. The findings support cautious, evidence-based use as a scoring support tool rather than a replacement for human judgement.

## 1. Introduction

Short-answer assessment is valuable because it can elicit forms of student understanding that are difficult to observe in selected-response formats. A short written response may reveal partial knowledge, missing concepts, misconceptions, causal or relational errors, and the degree to which a student can explain a scientific idea in their own words (Black & Wiliam, 1998; Burrows et al., 2015). These same features make short-answer scoring difficult. Student answers can be brief, lexically diverse, incomplete, or partly correct, and a scoring method must distinguish meaningful paraphrase from superficial word overlap, contradiction, irrelevance, and non-domain responses.

Automated short-answer scoring has often relied on lexical, semantic, or machine-learning approaches, including answer length, keyword overlap, TF-IDF cosine similarity, semantic similarity, and supervised classification (Burrows et al., 2015; Mohler et al., 2011; Salton & Buckley, 1988). These approaches provide useful baselines because they are transparent, reproducible, and often capture part of the signal in short responses. At the same time, surface similarity is an incomplete basis for diagnostic scoring. A response may share words with the reference answer while expressing an incorrect relation, or it may express the intended scientific meaning with little lexical overlap. For assessment purposes, the key issue is not only whether an automated score is accurate, but whether it provides defensible evidence about the quality of the response.

Large language models (LLMs) create new possibilities for short-answer scoring because they can compare a question, reference answer, and student answer in natural language and return structured judgements (Kasneci et al., 2023). Yet LLM scoring should not be treated as a deterministic single judgement. Many evaluations emphasize single-model, single-run, accuracy-focused performance. Such evidence is useful but incomplete for educational assessment. Score use also requires evidence about reliability, validity, scoring consistency, model severity, diagnostic value, and failure cases (American Educational Research Association et al., 2014; Messick, 1995; Williamson et al., 2012). Repeated-run analysis is especially relevant because LLM outputs may vary across generations, while multi-model analysis is relevant because different models may agree in ranking responses but differ in severity or error profile.

This study evaluates OCG-PRES guided LLM scoring as a reliability, validity, and diagnostic evidence problem rather than as a model ranking exercise. OCG-PRES is used as a structured five-dimensional framework for scoring short-answer quality: concept coverage, relation accuracy, reasoning completeness, contradiction control, and domain relevance. The framework is intended to make the basis for a score more explicit by separating whether a response includes relevant concepts, expresses correct relations, provides sufficient reasoning, avoids contradiction, and remains within the scientific domain.

The empirical design uses 996 SciEntsBank short-answer responses scored by GPT, DeepSeek, and Qianwen across three independent runs. Scores are evaluated against official binary labels, official five-category labels, disagreement and failure cases, and traditional non-LLM textual baselines. The study addresses four research questions:

RQ1: To what extent are OCG-PRES guided LLM scores stable across repeated runs within each model?

RQ2: To what extent do GPT, DeepSeek, and Qianwen agree in their scores, and do they differ in scoring severity?

RQ3: How well do OCG-PRES guided LLM scores align with official binary and five-category labels?

RQ4: Does OCG-PRES guided LLM scoring outperform traditional non-LLM textual baselines?

## 2. Literature Review / Background

### 2.1 Automated Short-Answer Scoring and Traditional Textual Baselines

Automated short-answer scoring has a long history in educational assessment and natural language processing (Burrows et al., 2015; Dzikovska et al., 2013; Haller et al., 2022). Traditional approaches often compare student answers with reference answers through lexical overlap, term weighting, semantic similarity, or supervised machine learning. Keyword overlap and Jaccard similarity provide simple measures of shared vocabulary. TF-IDF cosine similarity adds term weighting and document-level normalization, which can reduce the influence of common words while emphasizing more distinctive terms (Manning et al., 2008; Salton & Buckley, 1988). These methods are transparent and reproducible, making them useful as baseline comparisons.

The main limitation of surface-text baselines is that they do not directly evaluate the scientific relation expressed by a student answer. A response can overlap lexically with a reference answer while reversing the causal relation, omitting the explanatory mechanism, or introducing a contradiction. Conversely, a correct response may use different wording from the reference answer. Semantic similarity and supervised models can partly address these limitations, but they still require validation against the intended construct and use context (Burrows et al., 2015; Mohler et al., 2011).

### 2.2 LLMs in Educational Assessment

LLMs can compare student language with reference materials and produce structured scores or rationales (Kasneci et al., 2023). In educational assessment, this capability may support low-stakes diagnostic feedback, formative review, and teacher triage. It also raises concerns about consistency, opacity, prompt sensitivity, model updates, bias, and over-reliance on automated judgement (American Educational Research Association et al., 2014; Kasneci et al., 2023; Williamson et al., 2012).

For these reasons, LLM scoring should be evaluated as a measurement problem. A high performance value on one dataset does not establish reliability or validity for assessment use. Evidence is needed across repeated runs, models, scoring thresholds, official labels, diagnostic categories, and failure cases.

### 2.3 Reliability and Validity Evidence in Automated Scoring

Reliability evidence concerns score consistency under relevant sources of variation (American Educational Research Association et al., 2014; Koo & Li, 2016; Shrout & Fleiss, 1979). For LLM scoring, repeated generation is one such source. Inter-model agreement is another useful source of evidence because different models may rank responses similarly while differing in severity.

Validity evidence concerns whether scores support intended interpretations and uses (American Educational Research Association et al., 2014; Messick, 1995). In this study, official binary labels provide one reference for correctness alignment, while official five-category labels provide a diagnostic reference for correct, partially correct incomplete, contradictory, irrelevant, and non-domain responses. Correlations, ROC/AUC, classification metrics, and category-level score patterns are treated as complementary evidence rather than as a single definitive validation result.

### 2.4 OCG-PRES Guided Scoring

OCG-PRES guided scoring is treated in this study as a structured scoring framework, not simply as a prompt format. The framework decomposes short-answer quality into five diagnostically meaningful dimensions. Concept coverage concerns whether the student answer includes the required scientific concepts. Relation accuracy concerns whether the answer correctly explains relationships among those concepts. Reasoning completeness concerns whether the answer gives enough explanatory reasoning to support the response. Contradiction control concerns whether the answer avoids claims that conflict with the reference answer. Domain relevance concerns whether the answer remains relevant to the scientific domain of the question.

OCG-PRES is not proposed as a new reasoning machine or a general theory of computation. Rather, it is positioned as an assessment-oriented operationalization of principles from knowledge representation and reasoning, ontology evaluation, and semantic web knowledge modeling. In knowledge representation and reasoning, a knowledge-based system must represent relevant entities and concepts, preserve meaningful relations among them, support inference, avoid inconsistency, and operate within an intended domain of interpretation (Brachman & Levesque, 2004). Ontology engineering frameworks such as OntoClean likewise evaluate conceptual models by the structural commitments they make, including identity, unity, dependence, and rigidity (Guarino & Welty, 2002). The W3C semantic web stack treats knowledge as structured entities and relations that can be represented, queried, checked, and reasoned over through standards such as RDF, OWL, and SPARQL (World Wide Web Consortium, 2012, 2013, 2014). OCG-PRES translates these computational knowledge-system principles into a measurement framework for short-answer assessment.

These dimensions matter because short-answer quality is rarely reducible to word overlap or a single correct/incorrect decision. A response can mention appropriate concepts but connect them incorrectly. It can be relevant but incomplete. It can include a scientifically meaningful phrase while also making a contradictory claim. It can use vocabulary from the question while remaining off task or outside the domain. OCG-PRES is intended to make these distinctions more visible by asking the scoring model to attend to different forms of evidence in the response.

In the Methods, each dimension is scored on a 0-4 scale. The mean score is the average of the five dimension scores and ranges from 0 to 4. The total score is the sum of the five dimension scores and ranges from 0 to 20. The derived mean and total scores support overall performance analyses, while the five dimensions may support diagnostic interpretation. In this study, diagnostic value is evaluated cautiously through alignment with official binary labels and score patterns across official five-category labels. These analyses provide evidence about OCG-PRES guided scoring, but they do not establish full validation and do not imply that automated scoring can replace human judgement.

## 3. Methods

### 3.1 Dataset and Official Labels

The study used 996 short-answer responses from the SciEntsBank dataset (Dzikovska et al., 2013). Each record included a question, a reference answer, a student answer, an official label text category, and a binary correct/incorrect label. The official labels were used as reference labels for evaluating score alignment. They were not treated as independent teacher ratings collected for this study.

Two official label structures were used. The binary label distinguished correct from incorrect responses. The five-category label distinguished correct, partially_correct_incomplete, contradictory, irrelevant, and non_domain responses. The binary label supported classification and ROC/AUC analyses, while the five-category label supported diagnostic pattern analysis.

### 3.2 OCG-PRES Guided Scoring Framework

OCG-PRES guided scoring decomposes short-answer quality into five dimensions: concept coverage, relation accuracy, reasoning completeness, contradiction control, and domain relevance. Concept coverage evaluates whether the answer includes required scientific concepts. Relation accuracy evaluates whether the relationships among concepts are stated correctly. Reasoning completeness evaluates whether the explanation is sufficiently developed. Contradiction control evaluates whether the response avoids claims that conflict with the reference answer. Domain relevance evaluates whether the answer remains within the scientific domain of the question.

The five dimensions were therefore defined as cognitive reliability indicators for a student response treated as a small knowledge claim. Concept coverage evaluates whether the response contains necessary conceptual entities; relation accuracy evaluates whether it preserves the relations among those entities; reasoning completeness evaluates inferential support; contradiction control evaluates consistency with the reference answer; and domain relevance evaluates whether the response remains inside the intended semantic domain. In this sense, OCG-PRES does not claim to implement a new formal reasoner. It defines task-specific key performance indicators for the quality of a knowledge representation and operationalizes those indicators as algorithmically scorable dimensions for educational assessment.

Each dimension was scored on a 0-4 scale. The response-level mean score was calculated as the average of the five dimensions, giving a 0-4 scale. The total score was calculated as the sum of the five dimensions, giving a 0-20 scale. The mean score was used as the main response-level score in most analyses.

### 3.3 Multi-Model Repeated Scoring Procedure

Three LLMs were included: GPT, DeepSeek, and Qianwen. Each model scored the same 996 responses across three independent runs. The final long-format dataset contained 8,964 scoring records, corresponding to 996 responses x 3 models x 3 runs. For each model, response-level scores were summarized as the three-run mean score. An ensemble score was calculated as the average of the nine model-run mean scores.

The study did not include independent human teacher ratings. It also did not include a direct holistic LLM baseline. The results should therefore be interpreted as evidence about OCG-PRES guided LLM scoring relative to official labels and non-LLM textual baselines.

### 3.4 Non-LLM Baseline Construction

Four non-LLM textual baselines were constructed from the same response data. The length baseline used student answer length. The keyword overlap baseline used Jaccard similarity between cleaned reference and student answers (Jaccard, 1901). The TF-IDF baseline used cosine similarity between corresponding reference and student answers. The combined traditional baseline used logistic regression with student word count, length ratio, overlap ratio relative to the reference answer, Jaccard similarity, and TF-IDF cosine similarity.

These baselines were included to evaluate whether OCG-PRES guided scoring provided evidence beyond answer length and surface lexical similarity. No LLM API calls were used to construct the non-LLM baselines.

### 3.5 Data Preparation and Quality Checks

The final long-format dataset contained 8,964 records and 996 distinct responses. Data quality checks found no duplicated response-model-run rows, no missing response-model-run combinations, no missing official labels, no missing official binary labels, no missing score values, and no out-of-range OCG-PRES scores.

### 3.6 Analytical Strategy

RQ1 examined repeated-run stability within each model. Analyses included ICC(3,k), ICC(3,1), exact agreement rates, mean absolute differences, and within-model variability. ICC(3,k) summarized the reliability of averaged repeated scores, while ICC(3,1) summarized single-run consistency.

RQ2 examined inter-model agreement and scoring severity. Inter-model agreement was evaluated using pairwise Pearson and Spearman correlations among three-run mean scores and ICC treating the three models as raters. Severity differences were evaluated using repeated-measures ANOVA, Friedman tests, pairwise paired t-tests, Wilcoxon signed-rank tests, and Cohen’s dz.

RQ3 evaluated alignment with official labels. Binary alignment was examined using Pearson and Spearman correlations, ROC/AUC (Hanley & McNeil, 1982), logistic regression, confusion matrices, accuracy, Cohen’s kappa (Cohen, 1960), precision, recall, specificity, F1, and balanced accuracy. Fixed-threshold classification used mean_score >= 3.0. Diagnostic patterns were evaluated by comparing mean OCG-PRES scores across the five official label categories.

RQ4 compared OCG-PRES scores with non-LLM textual baselines. The baseline comparison used optimal thresholds selected within the comparison analysis. These optimal-threshold results are reported separately from the fixed threshold = 3.0 classification results.

## 4. Results

### 4.1 Data Quality and Descriptive Overview

The final long-format dataset contained 8,964 scoring records from 996 distinct responses, matching the expected design of 996 responses scored by three models across three repeated runs. Data quality checks found no duplicated response-model-run rows, no missing response-model-run combinations, no missing official labels, no missing official binary labels, no missing OCG-PRES score values, and no out-of-range OCG-PRES score values.

Mean scores on the 0-4 scale were 2.160 for GPT, 2.084 for DeepSeek, 1.986 for Qianwen, and 2.077 for the ensemble.

### 4.2 Repeated-Run Stability Within Each Model

All three models showed high repeated-run reliability based on response-level mean OCG-PRES scores (Table 1). GPT had ICC(3,k) = .977 and ICC(3,1) = .935. DeepSeek had ICC(3,k) = .992 and ICC(3,1) = .977. Qianwen had ICC(3,k) = .981 and ICC(3,1) = .945.

Item-level agreement indicators differed across models. DeepSeek had the highest average exact agreement (.722) and the lowest mean absolute difference (.099). GPT had average exact agreement of .361 and mean absolute difference of .242. Qianwen had average exact agreement of .410 and mean absolute difference of .211.

**Table 1**

*Repeated-run reliability of OCG-PRES scores*

| Model | ICC(3,k) | Exact agreement | Mean absolute difference |
|---|---|---|---|
| GPT | .977 | .361 | .242 |
| DeepSeek | .992 | .722 | .099 |
| Qianwen | .981 | .410 | .211 |

Note. ICC(3,k) summarizes the reliability of averaged repeated runs.

### 4.3 Inter-Model Agreement and Scoring Severity

Pairwise Pearson correlations among model-level three-run mean scores were high: GPT vs DeepSeek $r = .914$, GPT vs Qianwen $r = .900$, and DeepSeek vs Qianwen $r = .898$. The inter-model ICC was ICC(3) = .903 and ICC(3,k) = .965.

The models differed in mean score. GPT had the highest mean score (2.160), followed by DeepSeek (2.084) and Qianwen (1.986). The repeated-measures ANOVA indicated a model effect, $p < .001$. The Friedman test was also significant, $p < .001$. Pairwise comparisons showed that GPT scored higher than DeepSeek by .076 points, $p < .001$, $d_z = .185$; GPT scored higher than Qianwen by .175 points, $p < .001$, $d_z = .407$; and DeepSeek scored higher than Qianwen by .098 points, $p < .001$, $d_z = .218$.

**Table 2**

*Inter-model agreement and scoring severity*

| Component | Result |
|---|---|
| GPT vs DeepSeek Pearson r | .914 |
| GPT vs Qianwen Pearson r | .900 |
| DeepSeek vs Qianwen Pearson r | .898 |
| Inter-model ICC(3) | .903 |
| Inter-model ICC(3,k) | .965 |
| GPT mean score | 2.160 |
| DeepSeek mean score | 2.084 |
| Qianwen mean score | 1.986 |
| Ensemble mean score | 2.077 |
| GPT - DeepSeek | Difference = .076, $p < .001$, $d_z = .185$ |
| GPT - Qianwen | Difference = .175, $p < .001$, $d_z = .407$ |
| DeepSeek - Qianwen | Difference = .098, $p < .001$, $d_z = .218$ |

Note. Mean scores are on the 0-4 OCG-PRES mean-score scale.

### 4.4 Alignment With Official Binary Labels

OCG-PRES scores were positively associated with official_binary (Table 3). GPT had Pearson r = .626 and AUC = .909. DeepSeek had r = .596 and AUC = .900. Qianwen had r = .556 and AUC = .872. The ensemble had r = .612 and AUC = .903.

Using the fixed threshold mean_score >= 3.0, GPT achieved accuracy = .845, kappa = .589, precision = .692, recall = .692, and F1 = .692. DeepSeek achieved accuracy = .840, kappa = .583, precision = .673, recall = .708, and F1 = .690. Qianwen achieved accuracy = .830, kappa = .494, precision = .740, recall = .500, and F1 = .597. The ensemble achieved accuracy = .843, kappa = .554, precision = .735, recall = .588, and F1 = .653.

**Table 3**

*Official binary alignment and classification performance*

| Model | r | AUC | Accuracy | Kappa | Precision | Recall | F1 |
|---|---|---|---|---|---|---|---|
| GPT | .626 | .909 | .845 | .589 | .692 | .692 | .692 |
| DeepSeek | .596 | .900 | .840 | .583 | .673 | .708 | .690 |
| Qianwen | .556 | .872 | .830 | .494 | .740 | .500 | .597 |
| Ensemble | .612 | .903 | .843 | .554 | .735 | .588 | .653 |

Note. Classification metrics use the fixed OCG-PRES threshold mean_score >= 3.0.

### 4.5 Diagnostic Patterns Across Official Five-Category Labels

Mean OCG-PRES scores varied by official five-category label (Table 4). Correct responses received the highest scores across models: GPT = 3.178, DeepSeek = 3.133, Qianwen = 2.922, and ensemble = 3.078. Partially correct incomplete responses received lower scores: GPT = 2.533, DeepSeek = 2.515, Qianwen = 2.392, and ensemble = 2.480.

Irrelevant responses had means of 1.550 for GPT, 1.623 for DeepSeek, 1.532 for Qianwen, and 1.569 for the ensemble. Contradictory responses had means of 1.483 for GPT, 1.099 for DeepSeek, 1.222 for Qianwen, and 1.268 for the ensemble. Non-domain responses had the lowest means: .864 for GPT, .823 for DeepSeek, .471 for Qianwen, and .719 for the ensemble.

**Table 4**

*Mean OCG-PRES scores across official five-category labels*

| Official label | GPT | DeepSeek | Qianwen | Ensemble |
|---|---|---|---|---|
| Correct | 3.178 | 3.133 | 2.922 | 3.078 |
| Partially correct incomplete | 2.533 | 2.515 | 2.392 | 2.480 |
| Irrelevant | 1.550 | 1.623 | 1.532 | 1.569 |
| Contradictory | 1.483 | 1.099 | 1.222 | 1.268 |
| Non-domain | .864 | .823 | .471 | .719 |

Note. Scores are model-level three-run means on the 0-4 scale.

### 4.6 Comparison With Non-LLM Textual Baselines

Table 5 reports the optimal-threshold comparison between non-LLM baselines and OCG-PRES scores. The length baseline had AUC = .585, F1 = .424, accuracy = .434, and kappa = .081. The Jaccard baseline had AUC = .741, F1 = .527, accuracy = .694, and kappa = .317. The TF-IDF baseline had AUC = .710, F1 = .507, accuracy = .612, and kappa = .243. The combined traditional baseline had AUC = .756, F1 = .538, accuracy = .676, and kappa = .313.

All OCG-PRES methods exceeded the non-LLM baselines in AUC and F1. In the same optimal-threshold comparison, OCG-PRES GPT had AUC = .909 and F1 = .688, OCG-PRES DeepSeek had AUC = .900 and F1 = .698, OCG-PRES Qianwen had AUC = .872 and F1 = .648, and the OCG-PRES ensemble had AUC = .903 and F1 = .695. These values are not directly interchangeable with the fixed threshold = 3.0 results reported in Section 4.4.

**Table 5**

*Comparison between non-LLM baselines and OCG-PRES scores*

| Method | AUC | F1 | Accuracy | Kappa |
|---|---|---|---|---|
| Length baseline | .585 | .424 | .434 | .081 |
| Jaccard baseline | .741 | .527 | .694 | .317 |
| TF-IDF baseline | .710 | .507 | .612 | .243 |
| Combined traditional baseline | .756 | .538 | .676 | .313 |
| OCG-PRES GPT | .909 | .688 | .790 | .544 |
| OCG-PRES DeepSeek | .900 | .698 | .833 | .584 |
| OCG-PRES Qianwen | .872 | .648 | .764 | .487 |
| OCG-PRES Ensemble | .903 | .695 | .819 | .571 |

Note. This table uses optimal thresholds for the baseline comparison. These values should not be mixed with the fixed threshold = 3.0 results in Table 3.

## 4.7 Disagreement and Failure Cases

Under the fixed threshold mean_score >= 3.0, GPT produced 77 false positives and 77 false negatives. DeepSeek produced 86 false positives and 73 false negatives. Qianwen produced 44 false positives and 125 false negatives. The ensemble produced 53 false positives and 103 false negatives.

Across models, 89 cases were three-model consistent errors, and 155 cases showed three-model disagreement. The top 5% highest model-disagreement set contained 50 cases, and the top 10% set contained 100 cases.

**Table 6**

*Disagreement and failure case summary*

| Model / case type | False positives | False negatives | Notes |
|---|---|---|---|
| GPT | 77 | 77 | Balanced error counts under threshold = 3.0. |
| DeepSeek | 86 | 73 | Slightly more false positives than false negatives. |

| Model / case type | False positives | False negatives | Notes |
|---|---|---|---|
| Qianwen | 44 | 125 | Fewer false positives but substantially more false negatives. |
| Ensemble | 53 | 103 | Fewer false positives than GPT and DeepSeek, with more false negatives. |
| Three-model consistent errors | NA | NA | 89 cases. |
| Three-model disagreement cases | NA | NA | 155 cases. |

### 4.8 Supplementary Results Previously Planned as Appendices

The analyses originally planned as appendices were incorporated into the Results section as compact tables where possible. Table 7 summarizes dimension-level correlations with the official binary label. Table 8 summarizes threshold sensitivity by selecting the best F1 value within the tested thresholds for each model. Table 9 documents the remaining supplementary result files and reproducibility materials that support the reported analyses without placing additional tables at the end of the manuscript.

**Table 7**

*Dimension-level correlations with official binary labels*

| Model | Dimension | Pearson r | Spearman rho |
|---|---|---|---|
| GPT | Concept coverage | .606 | .584 |
| GPT | Relation accuracy | .651 | .621 |
| GPT | Reasoning completeness | .571 | .551 |
| GPT | Contradiction control | .430 | .480 |
| GPT | Domain relevance | .389 | .470 |
| DeepSeek | Concept coverage | .587 | .564 |
| DeepSeek | Relation accuracy | .605 | .606 |
| DeepSeek | Reasoning completeness | .598 | .574 |

| Model | Dimension | Pearson r | Spearman rho |
|---|---|---|---|
| DeepSeek | Contradiction control | .285 | .300 |
| DeepSeek | Domain relevance | .476 | .533 |
| Qianwen | Concept coverage | .493 | .488 |
| Qianwen | Relation accuracy | .590 | .566 |
| Qianwen | Reasoning completeness | .495 | .471 |
| Qianwen | Contradiction control | .340 | .360 |
| Qianwen | Domain relevance | .443 | .497 |

Note. Correlations use model-level three-run mean dimension scores and official_binary.

The dimension-level correlations showed that relation accuracy had the strongest Pearson association with official_binary for GPT (r = .651), DeepSeek (r = .605), and Qianwen (r = .590). Concept coverage and reasoning completeness also showed moderate to strong associations, while contradiction control and domain relevance varied more across models.

**Table 8**

*Threshold sensitivity summary by best F1 score*

| Model | Best threshold | Accuracy | Kappa | Precision | Recall | Specificity | F1 |
|---|---|---|---|---|---|---|---|
| GPT | 3.0 | .845 | .589 | .692 | .692 | .897 | .692 |
| DeepSeek | 3.0 | .840 | .583 | .673 | .708 | .885 | .690 |
| Qianwen | 2.5 | .798 | .523 | .571 | .788 | .802 | .662 |
| Ensemble | 2.5 | .791 | .529 | .555 | .852 | .771 | .672 |

Note. Best threshold is selected within the tested thresholds of 2.0, 2.5, 3.0, 3.5, and 4.0. These threshold-sensitivity values are supplementary to the fixed threshold = 3.0 results in Table 3.

The threshold-sensitivity analysis showed that the best F1 threshold was 3.0 for GPT and DeepSeek, but 2.5 for Qianwen and the ensemble. These results indicate that fixed-threshold performance and threshold-optimized performance answer different evaluation questions.

**Table 9**

*Supplementary results and reproducibility materials incorporated into the Results section*

| Former appendix item | Content | Project file or location | Use in manuscript |
|---|---|---|---|
| A | Full repeated-run correlations | outputs/table_04_repeated_run_correlations.csv | Supports RQ1 reliability interpretation. |
| B | Dimension-level official-binary correlations | outputs/ table_17_dimension_correlations_with_official_binary.csv | Summarized in Table 7. |
| C | Threshold sensitivity analysis | outputs/table_21_threshold_sensitivity.csv | Summarized in Table 8. |
| D | Logistic regression results | outputs/table_16_logistic_total_score_models.csv and outputs/table_18_dimension_logistic_regression.csv | Supports official-label alignment analyses. |
| E | Non-LLM baseline feature details | outputs/non_llm_baselines/ | Supports RQ4 baseline comparison. |
| F | False positive and false negative case files | outputs/cases_false_positives_*.csv and outputs/cases_false_negatives_*.csv | Supports failure-case review. |
| G | High-disagreement case files | outputs/cases_high_model_disagreement_top5.csv and outputs/cases_high_model_disagreement_top10.csv | Supports uncertainty and review analyses. |
| H | R scripts and reproducibility notes | analysis_ocg_pres_multimodel.R and non_llm_baseline_analysis.R | Supports reproducibility. |

Note. Case-level files should be checked against dataset-use permissions before any verbatim student responses are quoted in a published article.

## 5. Discussion

### 5.1 Repeated-Run Scoring as Reliability Evidence

The repeated-run analyses provide reliability evidence for OCG-PRES guided LLM scoring. All three models had high ICC(3,k) values, indicating that averaged scores were stable across runs. DeepSeek was the most stable model, with ICC(3,k) = .992, average exact agreement = .722, and mean absolute difference = .099. GPT and Qianwen also had high ICC values, but their exact agreement and MAD results showed more item-level variation.

This distinction is important for assessment use. ICC summarizes consistency in score ordering and averaged scores, whereas exact agreement and MAD show how much a score may vary for a particular response across repeated runs. A scoring system may therefore appear highly reliable in aggregate while still producing item-level variation that matters for individual cases. Repeated-run analysis makes this source of uncertainty visible.

### 5.2 Stability, Severity, and Official-Label Alignment as Distinct Properties

The results show that stability, severity, and alignment with official labels are related but distinct properties. DeepSeek was the most stable model, but GPT had the highest Pearson correlation with official_binary and the highest AUC. Qianwen was stricter, with higher precision but lower recall under the fixed threshold = 3.0 rule. These patterns suggest that model evaluation should not rely on a single metric.

Severity differences also matter for score interpretation. GPT assigned the highest mean scores, Qianwen the lowest, and DeepSeek fell between them. The models agreed strongly in their ranking of responses, yet they applied systematically different score levels. In applied assessment settings, such severity differences could affect classification decisions, feedback, and the number of responses flagged for review.

The ensemble did not clearly outperform the strongest individual model. This does not mean that ensemble scoring lacks value, but it indicates that ensemble benefits should be tested empirically rather than assumed.

### 5.3 Diagnostic Value of OCG-PRES

The five-category results provide diagnostic evidence for OCG-PRES guided scoring. Across models, correct responses received the highest scores, partially correct incomplete responses were lower, irrelevant and contradictory responses were lower again, and non-domain responses received the lowest scores. This pattern is consistent with the intended ordering of the official categories and suggests that the score structure reflected more than binary correctness.

Model-specific patterns were also informative. DeepSeek assigned particularly low scores to contradictory responses, while Qianwen assigned especially low scores to non-domain responses. These differences may reflect model-specific sensitivity to different error types. Such patterns are relevant for diagnostic assessment because the nature of an error can matter as much as the final correct/incorrect decision.

### 5.4 Evidence Beyond Surface Text Similarity

The comparison with non-LLM baselines provides evidence that OCG-PRES guided scoring captured information beyond answer length and lexical similarity. The length baseline performed weakly. Jaccard similarity and TF-IDF cosine similarity performed better, indicating that surface overlap contained useful signal. The combined traditional baseline was the strongest non-LLM method, but all OCG-PRES methods exceeded it in AUC and F1 in the optimal-threshold comparison.

This result is consistent with the rationale for OCG-PRES. Short-answer quality depends on whether the response includes relevant concepts, expresses correct relations, provides adequate reasoning, avoids contradiction, and remains domain relevant. These features are not fully represented by length or lexical overlap. The findings therefore support the value of evaluating structured diagnostic scoring against simple textual baselines.

### 5.5 Implications for LLM-Assisted Educational Assessment

For assessment research, the study illustrates an evaluation framework that combines repeated-run reliability, inter-model agreement, severity analysis, official-label alignment, diagnostic category patterns, non-LLM baseline comparison, and failure-case analysis. This broader evidence structure is more appropriate for educational scoring than a single accuracy estimate.

For practice, the findings support cautious use of LLM-assisted scoring in contexts where scores inform feedback, review, or triage rather than final high-stakes decisions. High-disagreement cases, consistent model errors, and borderline classifications should be treated as signals for human review. The results provide evidence about the behaviour of OCG-PRES guided scoring in this dataset; they do not establish that the method should replace teacher judgement.

### 5.6 Limitations and Future Research

Several limitations should guide interpretation. First, the study did not include independent human teacher ratings. Official labels were used as reference labels, but they may not capture all scoring nuance. Second, the study did not include a direct holistic LLM baseline, so it cannot determine whether OCG-PRES improves over unguided holistic LLM scoring. Third, the dataset was limited to SciEntsBank science short-answer responses. Results may differ across subjects, languages, age groups, item formats, and assessment contexts.

Fourth, model versions and API behaviour may change over time. The results describe the outputs analysed in this project rather than permanent properties of GPT, DeepSeek, or Qianwen. Fifth, the non-LLM baselines are useful but not exhaustive; other traditional, neural, or embedding-based baselines could be added in future work. Finally, the diagnostic interpretation of OCG-PRES should be treated as evidence for further validation, not as full validation.

Future research should add independent human teacher scoring, compare OCG-PRES with direct holistic LLM scoring, test the framework across additional datasets and domains, examine feedback quality, and evaluate human-AI moderation workflows. Future studies should also examine whether disagreement indicators can support uncertainty-aware scoring and targeted human review.

## 6. Conclusion

This study evaluated repeated multi-model OCG-PRES guided LLM scoring for 996 SciEntsBank short-answer responses. The results provide evidence of high repeated-run reliability, strong inter-model agreement, systematic severity differences, alignment with official binary labels, and diagnostic differentiation across official five-category labels. DeepSeek was the most stable across repeated runs, GPT showed the strongest official binary alignment, and Qianwen was the most stringent.

OCG-PRES guided scoring also outperformed traditional non-LLM baselines based on answer length, Jaccard similarity, TF-IDF cosine similarity, and a combined logistic model. These findings support further research on LLM-assisted diagnostic scoring while underscoring that such systems should support, not replace, human judgement.

## 7. Limitations and Future Research

The main limitations are summarized in Section 5.6. The study did not include independent human teacher ratings or a direct holistic LLM baseline, used one science short-answer dataset, and relied on official labels as reference labels. Future work should add teacher scoring, compare OCG-PRES with direct holistic LLM scoring, evaluate additional datasets and domains, and examine human-AI moderation workflows.

## 8. AI-Use Disclosure Statement

AI-assisted tools were used to support code organisation, manuscript outlining, and language editing. All analyses, numerical results, citations, and interpretations were checked and verified by the author.

## 9. Data and Code Availability Statement

The analysis scripts and derived result tables are available from the author upon reasonable request. The use and redistribution of the original dataset are subject to the dataset provider's terms and conditions.